\documentstyle[aclap,psfig]{article}
\newcommand {\deno} [1] {[\![#1]\!]}
\newcommand{\bsl}{\backslash}

\title{Efficient Construction of Underspecified Semantics under Massive Ambiguity}
\author{Jochen D\"orre\thanks{This research has been carried
    out while the author visited the Programming Systems Lab of Prof. Gert
    Smolka at the University of Saarland, Saarbr\"ucken. Thanks to
    John Maxwell, Martin M\"uller, Joachim Niehren, Michael Schiehlen,
    and an anonymous reviewer for valuable feedback and to all at PS
    Lab for their helpful support with the OZ system.}\\
  Institut f\"ur maschinelle Sprachverarbeitung\\
  University of Stuttgart}

\begin{document}
%\makeidpage
\maketitle

%\vspace*{-7ex}
\begin{abstract}
We investigate the problem of determining a compact
  underspecified semantical representation for sentences that may be
  highly ambiguous. Due to combinatorial explosion, the
  naive method of building semantics for the different syntactic readings
  independently is prohibitive. We
  present a method that takes as input a syntactic parse forest with
  associated constraint-based semantic construction rules and directly
  builds a {\em packed semantic structure}. The algorithm is fully implemented
  and runs in $O(n^4 log(n))$ in sentence length, if the grammar
  meets some reasonable `normality' restrictions.
\end{abstract}

\section{Background}

One of the most central problems that any NL system must face is the
ubiquitous phenomenon of ambiguity. In the last few years a whole new
branch developed in semantics that investigates {\em underspecified}
semantic representations in order to cope with this phenomenon.
Such representations do not stand for the real or intended meaning of
sentences, but rather for the possible options of interpretation.
Quantifier scope ambiguities are a semantic variety of ambiguity that
is handled especially well by this approach. Pioneering work in that
direction has been \cite{Alshawi:1992} and \cite{Reyle93:UDRS}.

More recently there has been growing interest in developing the
underspecification approach to also cover syntactic ambiguities (cf.
\cite{Pinkal95,EggLebeth95,Schiehlen96}). Schiehlen's approach is
outstanding in that he fully takes into account syntactic constraints.
In \cite{Schiehlen96} he presents an algorithm which directly
constructs a single underspecified semantic structure from the ideal
``underspecified'' syntactic structure, a parse forest. 

On the other hand, a method for producing ``packed semantic
structures'', in that case ``packed quasi-logical forms'', has already
been used in the Core Language Engine, informally described in
\cite[Chap.~7]{Alshawi:1992}. However, this method only produces a
structure that is virtually isomorphic to the parse forest, since it
simply replaces parse forest nodes by their corresponding semantic
operators. No attempt is made to actually apply semantic operators in
the phase where those ``packed QLFs'' are constructed.
Moreover, the packing of the QLFs seems to serve no purpose in the
processing phases following semantic analysis. Already the immediately
succeeding phase ``sortal filtering'' requires QLFs to be unpacked,
i.e. enumerated.

Contrary to the CLE method, Schiehlen's method actively packs semantic
structures, even when they result from distinct syntactic structures,
extracting common parts.
His method, however, may take time exponential w.r.t. sentence
length. Already the semantic representations it produces can be
exponentially large, because they grow linear with the number of
(syntactic) readings and that can be exponential, e.g., for sentences
that exhibit the well-known attachment ambiguity of prepositional
phrases. It is therefore an interesting question to ask, whether we
can compute compact semantic representations from parse forests
without falling prey to exponential explosion. 

The purpose of the present paper is to show that construction of
compact semantic representations like in Schiehlen's
approach from parse forests is not only possible, but also cheap,
i.e., can be done in polynomial time.

\begin{figure*}[htb]
\mbox{\psfig{file=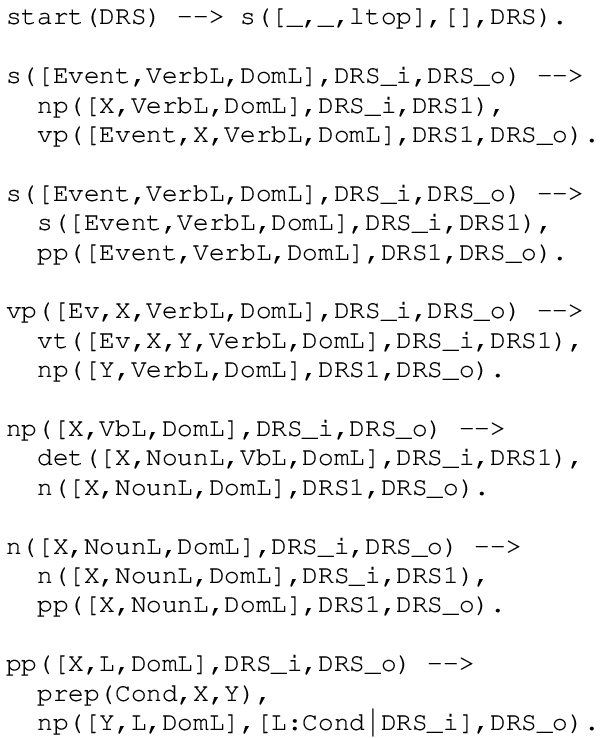}}
\hfill
\mbox{\psfig{file=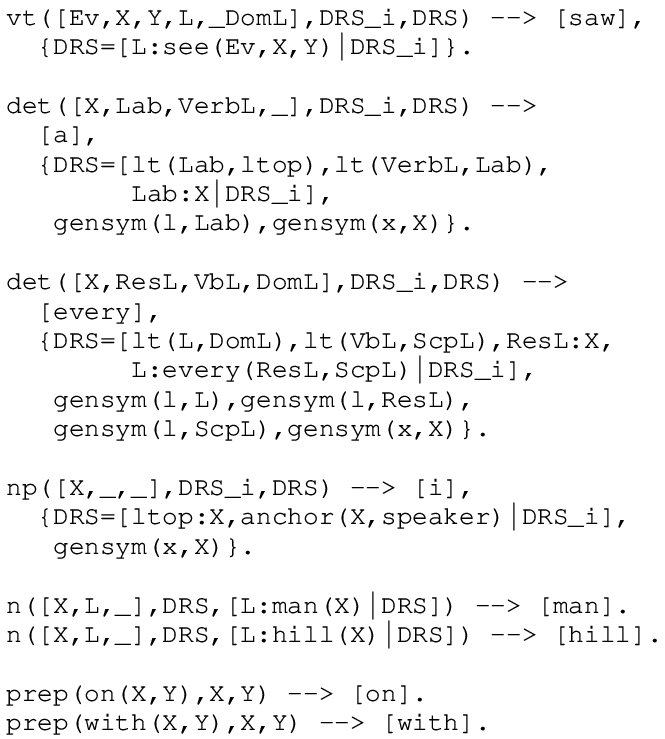}}
\caption{Example DCG}
\label{fig:pp-att-dcg}
\end{figure*}
To illustrate our method we use a simple DCG grammar for PP-attachment
ambiguities, adapted from \cite{Schiehlen96}, that yields semantic
representations (called UDRSs) according to the Underspecified
Discourse Representation Theory \cite{Reyle93:UDRS,Kamp/Reyle:1993}. The
grammar is shown in Fig.~\ref{fig:pp-att-dcg}.

The UDRSs constructed by the grammar are flat lists of the
UDRS-constraints $l\leq l'$ (subordination (partial) ordering between
labels; Prolog representation: {\tt lt($l$,$l'$)}),
$l:Cond$ (condition introduction in subUDRS labeled $l$), $l:X$
(referent introduction in $l$), $l:GenQuant(l',l'')$ (generalised
quantifier) and an anchoring function. The meaning of a UDRS as a set
of denoted DRSs can be explained as follows.%
\footnote{Readers unfamiliar with DRT should
  think of these structures as some Prolog terms, representing
  semantics, built by unifications according to the semantic rules. It
  is only important to notice how we extract common parts of those
  structures, irrespective of the structures' meanings.
 }
All conditions with the same label form a subUDRS and labels occurring
in subUDRSs denote locations (holes) where other subUDRSs can be
plugged into. The whole UDRS denotes the set of well-formed DRSs that
can be formed by some plugging of the subUDRSs that does not violate
the ordering $<$. Scope of quantifiers can be underspecified in
UDRSs, because subordination can be left partial.

In our example grammar every nonterminal has three arguments. The 2nd
and the 3rd argument represent a UDRS list as a difference list, i.e.,
the UDRS is ``threaded through''. The first argument is a list of
objects occurring in the UDRS that play a specific
role in syntactic combinations of the current node.%
\footnote{E.g., for an NP its referent, as well as the upper and lower
label for the current clause and the top label.}
  An example of a
UDRS, however a packed UDRS, is shown later on in \S\ref{sec:impl}.

To avoid the dependence on a particular grammar formalism we
present our method for a constraint-based grammar abstractly from the
actual constraint system employed. We only require that semantic rules
relate the semantic `objects' or structures that are associated with
the nodes of a local tree by employing constraints. E.g., we
can view the DCG rule $s \longrightarrow np\; vp$ as a relation between
three `semantic construction terms' or variables {\tt SemS, SemNP,
  SemVP} equivalent to the constraints\\[-3.5ex]
{\small
\begin{verbatim}
SemS = [[Event,VerbL,DomL,TopL],DRS_i,DRS_o]
SemNP = [[X,VerbL,DomL,TopL],DRS_i,DRS1]
SemVP = [[Event,X,VerbL,DomL,TopL],DRS1,DRS_o]
\end{verbatim}
}

Here is an overview of the paper. \S\ref{sec:problem} gives
the preliminaries and assumptions needed to precisely state the
problem we want to solve. \S\ref{sec:algo} presents the abstract
algorithm. Complexity considerations follow in
\S\ref{sec:complexity}. Finally, we consider implementation
issues, present results of an experiment in \S\ref{sec:impl},
and close with a discussion.

\section{The Problem}
\label{sec:problem}

As mentioned already, we aim at calculating from given parse forests
the same compact semantic structures that have been proposed by 
\cite{Schiehlen96}, i.e. structures that make explicit the common
parts of different syntactic readings, so that subsequent semantic
processes can use this generalised information. As he
does, we assume a constraint-based grammar, e.g. a DCG \cite{DCG80} or
HPSG \cite{Pollard/Sag:1994} ,
in which syntactic constraints and constraints that determine a
resulting semantic representation can be seperated and parsing can be
performed using the syntactic constraints only. 

\begin{figure*}[htb]
\centerline{\psfig{file=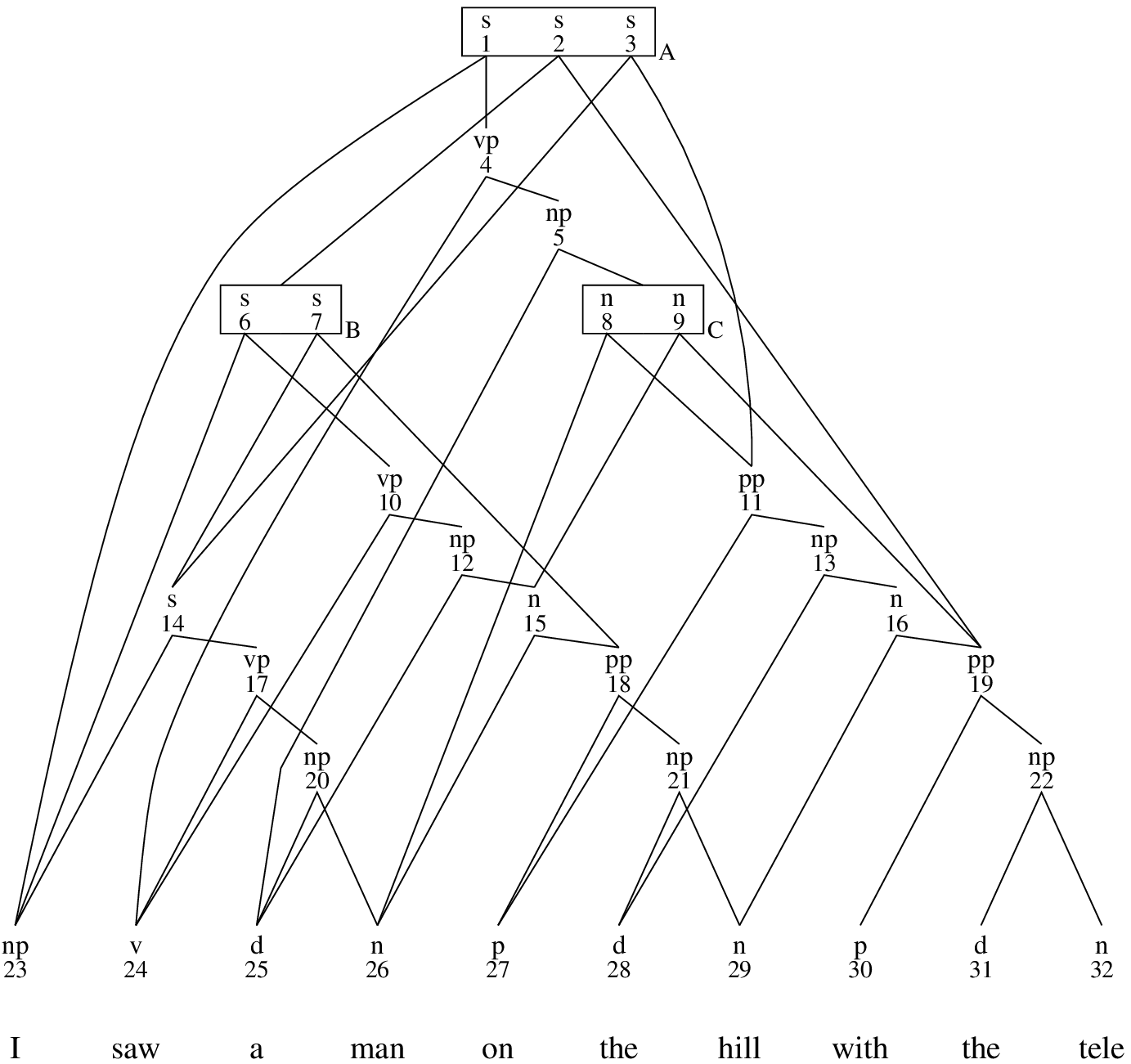}}
\caption{Example of a parse forest}
\label{fig:forest1}
\end{figure*}

Second, we assume that
the set of syntax trees can be compactly represented as a parse forest
(cf. \cite{Earley70,BillotLang89,Tomita86}). Parse forests are
rooted labeled directed acyclic graphs with {\sc AND}-nodes (standing
for context-free branching) and {\sc OR}-nodes (standing for
alternative subtrees), that can be characterised as follows (cf.
Fig.~\ref{fig:forest1} for an example).%
\footnote{The graphical representation of an {\sc OR}-node is a box surrounding its children, i.e. the {\sc
  AND-OR}-graph structure of 
\unitlength=1pt
\begin{picture}(36,15)(0,12)
\put(5,9){\framebox(26,15){S\hspace{7pt}S}}
\put(12,9){\line(-1,-1){10}}
\put(12,9){\line(1,-1){10}}
\put(24,9){\line(-1,-1){10}}
\put(24,9){\line(1,-1){10}}
\end{picture}
 is
\begin{picture}(54,15)(0,24)
\put(27,31.5){\circle{9}}
\put(31.5,27){\makebox(0,0)[l]{{\tiny OR}}}
\put(27,27){\line(-2,-1){18}}
\put(27,27){\line(2,-1){18}}
\put(9,13.5){\circle{9}}
\put(13.5,9){\makebox(0,0)[l]{{\tiny AND}}}
\put(9,13.5){\makebox(0,0){S}}
\put(45,13.5){\circle{9}}
\put(49.5,9){\makebox(0,0)[l]{{\tiny AND}}}
\put(45,13.5){\makebox(0,0){S}}
\put(9,9){\line(-1,-1){10}}
\put(9,9){\line(1,-1){10}}
\put(45,9){\line(-1,-1){10}}
\put(45,9){\line(1,-1){10}}
\end{picture}
.}
\begin{enumerate}
\item The terminal yield as well as the label of two {\sc AND}-nodes
  are identical, if and only if they both are children of one {\sc
  OR}-node. 
\item Every tree reading is a valid parse tree.
\end{enumerate}
Tree readings of such graphs are obtained by replacing any {\sc
  OR}-node by one of its children.
Parse forests can
represent an exponential number of phrase structure alternatives in
$o(n^3)$ space, where $n$ is the length of the sentence. The example
  uses the 3 {\sc OR}-nodes (A, B, C) and the
  {\sc AND}-nodes 1 through 32 to represent 5 complete parse trees,
  that would use $5\times 19$ nodes.

Third, we assume the rule-to-rule hypothesis, i.e., that the grammar
associates with each local tree a `semantic rule' that specifies how
to construct the mother node's semantics from those of its children.

Hence, input to the algorithm is
\begin{itemize}
\item a parse forest
\item an associated semantic rule for every local tree ({\sc AND}-node
  together with its children) therein
\item and a semantic representation for each leaf (coming from a
  semantic lexicon).
\end{itemize}

To be more precise, we assume a constraint language $\cal C$ over a
denumerable set of variables $\cal X$, that is a sublanguage of
Predicate Logic with equality and is closed
under conjunction, disjunction, and variable renaming. Small greek
letters $\phi, \psi$ will henceforth denote constraints (open
formulae) and letters 
$X, Y, Z$ (possibly with indeces) will denote variables. Writing
$\phi(X_1,\ldots,X_k)$ shall indicate that
$X_1,\ldots,X_k$ are the free variables in the constraint $\phi$.
Frequently used examples for constraint languages are 
 the language of equations over first-order terms for DCGs,%
\footnote{DCG shall refer in this paper to a logically pure version,
  Definite Clause Grammars based on pure {\sc Prolog}, involving no
  nonlogical devices like Cut, var/1, etc.}
 PATR-style feature-path equations, or
 typed feature structure description languages (like the
  constraint languages of ALE \cite{ALE} or CUF \cite{cuf2-dyana}) for
  HPSG-style grammars.

Together with the constraint language we require a constraint solver,
that checks constraints for satisfiability, usually by transforming
them into a normal form (also called `solved form'). Constraint
solving in the DCG case is simply unification of terms.

The semantic representations mentioned before are actually not given
directly, but rather as a constraint on some variable, thus allowing
for partiality in the structural description. To that end we assume
that every node in the parse forest $\nu$ has associated with it a
variable $X_{\nu}$ that is used for constraining the (partial)
semantic structure of $\nu$. The semantics of a leaf node $\mu$ is
hence given as a constraint $\phi_{\mu}(X_{\mu})$, called a {\em leaf 
  constraint}. 

A final assumption that we adopt concerns the nature of the `semantic
rules'. The process of semantics construction shall be a completely
monotonous process of gathering constraints that {\em never leads to
failure}. We assume that any
associated (instantiated) semantic rule $r(\nu)$ of a local tree ({\sc
  AND}-branching)
$\nu(\nu_1,\ldots,\nu_k)$ determines $\nu$'s semantics $\Sigma(\nu)$ as
follows from those of its children: 
$$ \Sigma(\nu) = \exists X_{\nu_1}\ldots\exists X_{\nu_k}
(
\begin{array}[t]{@{}l@{}}
\phi_{r(\nu)}(X_{\nu},X_{\nu_1},\ldots,X_{\nu_k})
\wedge\mbox{}\\ 
\Sigma(\nu_1) \wedge \ldots \wedge \Sigma(\nu_k)). 
\end{array}
$$
The constraint $\phi_{r(\nu)}(X_{\nu},X_{\nu_1},\ldots,X_{\nu_k})$ is
called the {\em rule constraint} for $\nu$. It is required to
only depend on the variables $X_{\nu},X_{\nu_1},\ldots,X_{\nu_k}$.
Note that if the same rule is to be applied at another node, we have a
different rule constraint.

Note that any $\Sigma(\nu)$ depends only on $X_{\nu}$ and can be
thought of as a unary predicate. Now, let us consider semantics
construction for a single parse tree for the moment. The leaf
constraints together with the rules define a semantics constraint
$\Sigma(\nu)$ for every node $\nu$, and the semantics of the full
sentence is described by the $\Sigma$-constraint of the root node,
$\Sigma(root)$. In the $\Sigma$-constraints, we actually can suppress 
the existential quantifiers by adopting the convention that any
variable other than the one of the current node is implicitly
existentially bound on the formula toplevel. Name conflicts, that
would force variable renaming, cannot occur. Therefore $\Sigma(root)$
is (equivalent to) just a big conjunction of all rule constraints for
the inner nodes and all leaf constraints.

Moving to parse forests, the semantics of an {\sc OR}-node
$\nu(\nu_1,\ldots,\nu_k)$ is to be defined as
$$ \Sigma(\nu)=\exists X_{\nu_1}\ldots\exists X_{\nu_k}
(
\begin{array}[t]{@{}l@{}}
\Sigma(\nu_1)\wedge X_{\nu}{=}X_{\nu_1} \vee \ldots\\ 
\mbox{}\vee
\Sigma(\nu_k)\wedge X_{\nu}{=}X_{\nu_k}),
\end{array}
$$
specifying that the set of possible (partial) semantic representations
for $\nu$ is the union of those of $\nu$'s children.
However, we can simplify this formula once and for all by assuming
that for every {\sc OR}-node there is only one variable $X_{\nu}$ that
is associated with it {\em and all of its children}. Using the same
variable for $\nu_1 \ldots \nu_k$ is unproblematic, because no two of
these nodes can ever occur in a tree reading. Hence, the definition we
get is
$$ \Sigma(\nu)=\Sigma(\nu_1) \vee \ldots \vee
\Sigma(\nu_k).
$$
Now, in the same
way as in the single-tree case, we can directly ``read off'' the
$\Sigma$-constraint for the whole parse forest representing the
semantics of all readings. Although this constraint is only half the
way to the packed semantic representation we are aiming at, it is
nevertheless worthwhile to consider its structure a little more
closely. Fig.~\ref{fig:sigma-constr} shows the structure of the
$\Sigma$-constraint for the {\sc OR}-node $B$ in  the example parse
forest. 

\begin{figure*}[htb]
$$
\begin{array}{l}
\begin{array}[t]{@{}c@{}}
\underbrace{
\phi_{r(6)}\wedge\phi_{23}\wedge\phi_{r(10)}\wedge\phi_{24} \wedge 
\phi_{r(12)}\wedge\phi_{25}\wedge\phi_{r(15)}\wedge\phi_{26}\wedge
\begin{array}[t]{@{}c@{}}
\underbrace{\phi_{r(18)}\wedge\phi_{27}\wedge\phi_{r(21)}\wedge
  \phi_{28}\wedge \phi_{29}}\\
\Sigma(18)
\end{array}
}\\
\Sigma(6)
\end{array}\\
\vee\\
\begin{array}[t]{@{}c@{}}
\underbrace{
\phi_{r(7)}\wedge\phi_{r(14)}\wedge\phi_{23}\wedge\phi_{r(17)}\wedge\phi_{24}
\wedge\phi_{r(20)}\wedge\phi_{25}\wedge\phi_{26}\wedge
\begin{array}[t]{@{}c@{}}
\underbrace{\phi_{r(18)}\wedge\phi_{27}\wedge\phi_{r(21)}\wedge
  \phi_{28}\wedge \phi_{29}}\\
\Sigma(18)
\end{array}}\\
\Sigma(7)
\end{array}
\end{array}
$$
\caption{Constraint $\Sigma(B)$ of example parse forest}
\label{fig:sigma-constr}
\end{figure*}

In a way the structure of this constraint directly mirrors the
structure of the parse forest. However, by writing out
the constraint, we loose the sharings present in the forest.
A subformula coming from a shared subtree (as $\Sigma(18)$ in
Fig.~\ref{fig:sigma-constr}) has to be stated as
many times as the subtree appears in an unfolding of the forest graph.
In our PP-attachment example the blowup caused by this is in fact
exponential. 

On the other hand, looking at a $\Sigma$-constraint as a
piece of syntax, we can {\em represent} this piece of syntax in the
same manner in which trees are represented in the parse forest, i.e.
we can have a representation of $\Sigma(root)$ with a structure
isomorphic to the forest's graph structure.%
\footnote{The packed QLFs in the Core Language
  Engine \cite{Alshawi:1992} are an example of such a representation.}
 In practice this
difference becomes a question of whether we have full control over the
representations the constraint solver employs (or any other process
that receives this constraint as input). If not, we cannot contend
ourselves with the {\em possibility} of compact representation of
constraints, but rather need a means to enforce this compactness on
the constraint level. This means that we have to introduce some form
of functional abstraction into the constraint language (or anything
equivalent that allows giving names to complex constraints and
referencing to them via their names). Therefore we enhance the
constraint language as follows. We allow to our disposition a second
set of variables, called names, and two special forms of constraints
\begin{quote}
\begin{tabular}{l@{\hspace{14ex}}r}
\multicolumn{2}{l}{1. {\tt def(<name>, <constraint>)}}\\
 & name definition\\
2. {\tt <name>} & name use
\end{tabular}
\end{quote}
with the requirements, that a name may only be used, if it is defined
and that its definition is unique. Thus, the constraint $\Sigma(B)$
above can be written as
$$\begin{array}{l}
(\begin{array}[t]{@{}l@{}}
\phi_{r(6)}\wedge\ldots\wedge\phi_{26}\wedge
\mbox{ \bf N }\\
\mbox{}\vee
\phi_{r(7)}\wedge\ldots\wedge\phi_{26}\wedge
\mbox{ \bf N })
\end{array}
\\
\mbox{}\wedge \mbox{ \tt def({\bf N},
  }\phi_{r(18)}\wedge\phi_{27}\wedge\phi_{r(21)}\wedge 
  \phi_{28}\wedge \phi_{29}\mbox{)}
\end{array}
$$

The packed semantic representation as constructed by the method
described so far still calls for an obvious improvement. Very often
the different branches of disjunctions contain constraints that have
large parts in common. However, although these overlaps are
efficiently handled on the representational level, they are invisible
at the {\em logical level}.
Hence, what we need is an algorithm that factores out common parts of
the constraints on the logical level, pushing disjunctions down.%
\footnote{Actually, in the $\Sigma(B)$ example such a factoring makes
 the use of the name {\bf N} superfluous. In 
 general, however, use of names is actually necessary to avoid
 exponentially large constraints. Subtrees may be shared by quite
 different parts of the structure, not only by disjuncts of the same
 disjunction. In the PP-attachment example, a compression of the
 $\Sigma$-constraint to polynomial size cannot be achieved with
 factoring alone.}
There are two routes that we can take to do this efficiently. 

In the first we consider only the structure of the parse forest,
however ignore the content of (rule or leaf) constraints. I.e. we
explore the fact 
that the parts of the $\Sigma$-constraints in a disjunction that stem
from nodes shared by all disjuncts must be identical, and hence can be
factored out.%
\footnote{\cite{MaxwellKaplan93} exploit the same idea for efficiently
  solving the functional constraints that an LFG grammar associates with a
  parse forest.}
More precisely, we can compute for every node $\nu$ the
set {\tt must-occur$(\nu)$} of nodes (transitively) dominated by $\nu$
that must occur in a tree of the forest, whenever $\nu$ occurs. We can
then use this information, when building the disjunction
$\Sigma(\nu)$ to factor
out the constraints introduced by nodes in {\tt must-occur$(\nu)$},
i.e., we build the factor $\Phi=\bigwedge_{v'\in\mbox{\tt
    must-occur}(\nu)} \Sigma(\nu')$ and a `remainder' constraint
$\Sigma(\nu_i)\bsl\Phi$ for each disjunct.

The other route goes one step further and takes into account the
content of rule and leaf constraints. For it we need an operation
{\tt generalise} that can be characterised informally as follows. 
\begin{quote}
For two satisfiable constraints $\phi$ and $\psi$, {\tt
  generalise}$(\phi,\psi)$ 
yields the triple $\xi,\phi',\psi'$, such that $\xi$ contains the
`common part' of $\phi$ and $\psi$ and $\phi'$ represents the
`remainder' $\phi\bsl\xi$ and likewise $\psi'$ represents
$\psi\bsl\xi$. 
\end{quote}

The exact definition of what the `common part' or the `remainder'
shall be,
naturally depends on the actual constraint system chosen. For our
purpose it is sufficient to require the following properties:
\begin{quote}
If {\tt generalise}$(\phi,\psi)\mapsto (\xi,\phi',\psi')$, then 
$\phi\vdash\xi$ and $\psi\vdash\xi$ and $\phi\equiv\xi\wedge\phi'$ and
$\psi\equiv\xi\wedge\psi'$.
\end{quote}

We shall call such a generalisation operation {\em simplifying} if the
normal form of $\xi$ is not larger than any of the input constraints'
normal form.

{\bf Example:} An example for such a generalisation operation for {\sc
  Prolog}'s constraint system (equations over first-order terms) is
  the so-called anti-unify operation, the dual of unification, that
  some {\sc Prolog} implementations provide as a library predicate.%
\footnote{{\tt anti\_unify} in Quintus Prolog
% \cite{quintusref}
, {\tt term\_subsumer} in Sicstus Prolog.}
% \cite{sicstusref}.}
  Two terms $T1$ and $T2$ `anti-unify' to $T$, iff $T$ is the (unique)
  most specific term that subsumes both $T1$ and $T2$. The `remainder
  constraints' in this case are the residual substitutions $\sigma_1$
  and $\sigma_2$ that transform $T$ into $T1$ or $T2$, respectively.

Let us now state the method informally. We use {\tt generalise} to factor
out the common parts of disjunctions. This is, however, not as trivial
as it might appear at first sight. {\tt Generalise} should operate on
solved forms, but when we try to eliminate the names introduced for
subtree constraints in order to solve the corresponding constraints,
we end up with constraints that are exponential 
in size. In the following section we describe an algorithm that
circumvents this problem.

\section{The Algorithm}
\label{sec:algo}

We call an order $\leq$ on the nodes of a directed acyclic graph
$G=(N,E)$ with nodes $N$ and edges $E$ {\em bottom-up}, iff whenever
$(i,j)\in E$ (``$i$ is a predecessor to $j$''), then $j\leq i$.

For the sake of simplicity let us assume that any nonterminal node in
the parse forest is binary branching. Furthermore, we leave implicit,
when conjunctions of constraints are normalised by the constraint
solver. Recall that for the generalisation operation it is usually
meaningful to operate on solved forms. However, at least the
simplifications $\mbox{\bf true}\wedge\phi\equiv\phi$ and
$\phi\wedge \mbox{\bf true}\equiv\phi$ should be assumed.

\newcommand{\SEM}{\mbox{\sc SEM}}
\newcommand{\DE}{\mbox{\sc D}}
\newcommand{\ENV}{\mbox{\sc ENV}}
\newlength{\myind}
\settowidth{\myind}{\mbox{else }}
\begin{figure*}[htb]
\begin{description}
\item[Input:] 
{\sf
\begin{tabular}[t]{l@{\hspace{2ex}}l}
$\bullet$ & parse forest, leaf and rule constraints as described
  above\\
$\bullet$ & array of variables $X_{\nu}$ indexed by
node s.t. if $\nu$ is a child of {\sc OR}-node $\nu'$, then
$X_{\nu}=X_{\nu'}$
% \newlength{\mywid}
% \setlength{\mywid}{\textwidth}
% \addtolength{\mywid}{-5cm}
% \begin{tabular}{@{}p{\mywid}@{}}
% an array containing the variables $X_{\nu}$ indexed by
%  node s.t. if $\nu$ is a child of an {\sc OR}-node $\nu'$, then
%  $X_{\nu}=X_{\nu'}$. 
% \end{tabular}

\end{tabular}
\item[Data structures:]
\begin{tabular}[t]{l@{\hspace{2ex}}l}
$\bullet$ & an array \SEM{} of constraints and an array \DE{} of
  names, both indexed by node\\
$\bullet$ & a stack \ENV{} of {\tt def} constraints 
\end{tabular}
\item[Output:] a constraint representing a packed semantic representation
\item[Method:]
\begin{tabular}[t]{l}
  \ENV{} := nil\\
  process nodes in a bottom-up order\\
  doing with node $\nu$:\\
  if $\nu$ is a leaf then\\
\hspace{\myind}\SEM{}[$\nu$] := $\phi_{\nu}$\\
\hspace{\myind}\DE{}[$\nu$] := {\bf true}\\
  elseif $\nu$ is AND($\nu_1,\nu_2$) then\\
\hspace{\myind}\SEM{}[$\nu$] := $\phi_{r(\nu)}\wedge \SEM{}[\nu_1]\wedge
  \SEM{}[\nu_2]$\\
\hspace{\myind}if \DE{}[$\nu_1$] = {\bf true} then \DE{}[$\nu$] :=
  \DE{}[$\nu_2$]\\
\hspace{\myind}elseif \DE{}[$\nu_2$] = {\bf true} then \DE{}[$\nu$] :=
  \DE{}[$\nu_1$]\\
\hspace{\myind}else \DE{}[$\nu$] := newname\\
\hspace{2\myind}push def(\DE{}[$\nu$], $\DE{}[\nu_1]\wedge\DE{}[\nu_2]$)
  onto \ENV\\
\hspace{\myind}end\\
  elseif $\nu$ is OR($\nu_1,\nu_2$) then\\
\hspace{\myind}let GEN, REM1, REM2 such that\\
\hspace{2\myind}generalise$(\SEM{}[\nu_1], \SEM{}[\nu_2])\mapsto
(\mbox{GEN},\mbox{REM1},\mbox{REM2})$\\
\hspace{\myind}\SEM{}[$\nu$] := GEN\\
\hspace{\myind}\DE{}[$\nu$] := newname\\
\hspace{\myind}push def(\DE{}[$\nu$], $\mbox{REM1}\wedge
  D[\nu_1]\vee\mbox{REM2}\wedge D[\nu_2]$) 
  onto \ENV\\
end
return $\SEM{}[root]\wedge \DE{}[root]\wedge \ENV{}$
\end{tabular}
}
\end{description}
\caption{Packed Semantics Construction Algorithm}
\label{fig:alg}
\end{figure*}

The Packed Semantics Construction Algorithm is given in
Fig.~\ref{fig:alg}. 
It enforces the following invariants, which can easily be shown by
induction.  
\begin{enumerate}
\item Every name used has a unique definition.
\item For any node $\nu$ we have the equivalence $\Sigma(\nu)\equiv \SEM{}[\nu]\wedge \deno{\DE{}[\nu]}$,
where $\deno{\DE{}[\nu]}$ shall denote the constraint obtained from
$\DE{}[\nu]$ when recursively replacing names by the constraints they
are bound to in \ENV{}.
\item For any node $\nu$ the constraint $\SEM{}[\nu]$ is never larger
  than the $\Sigma$-constraint of any single tree in the forest
  originating in $\nu$. 
\end{enumerate}
Hence, the returned constraint correctly represents the semantic
representation for all readings.

\section{Complexity}
\label{sec:complexity}

The complexity of this abstract algorithm depends primarily on the
actual constraint system and generalisation operation employed. But
note also that the influence of the actual semantic operations
prescribed by the grammar can be vast, even for the simplest
constraint systems. E.g., we can write a DCGs that produce abnormal
large ``semantic structures'' of sizes growing exponentially with
sentence length (for a single reading). For meaningful grammars we
expect this size function to be linear. Therefore, let us abstract
away this size by employing a function $f_G(n)$ that bounds the size
of semantic structures (respectively the size of its describing
constraint system in normal form) that grammar $G$ assigns to
sentences of length $n$. 

Finally, we want to assume that generalisation is simplifying and can be
performed within a bound of $g(m)$ steps, where $m$ is the total size
of the input constraint systems. 

With these assumptions in place, the time complexity for the algorithm
can be estimated to be ($n$ = sentence length, $N$ = number of forest
nodes) 
$$ O(g(f_G(n))\cdot N) \leq O(g(f_G(n))\cdot n^3),
$$
since every program step other than the generalisation operation can
be done in constant time per node. Observe that because of
Invariant~3.\ the input constraints 
to {\sf generalise} are bounded by $f_G$ as any constraint in \SEM{}.

In the case of a DCG the generalisation operation is {\tt anti\_unify},
which can be performed in $o(n\cdot log(n))$ time and space (for
acyclic structures). Hence,
together with the assumption that the semantic structures the DCG
computes can be bounded linearly in sentence length (and are acyclic),
we obtain a $O(n\cdot log(n)\cdot N) \leq O(n^4 log(n))$ total
time complexity. 
\begin{figure*}[t]
\begin{minipage}{.45\textwidth}
SEM[top]:

\centerline{\psfig{file=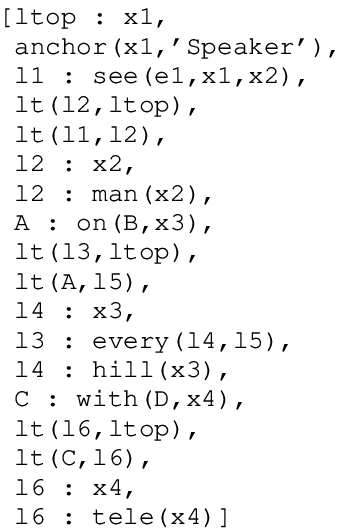}}

D[top] (a Prolog goal):

\centerline{\psfig{file=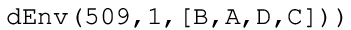}}
\end{minipage}
\begin{minipage}{.5\textwidth}
ENV (as Prolog predicates):

\centerline{\psfig{file=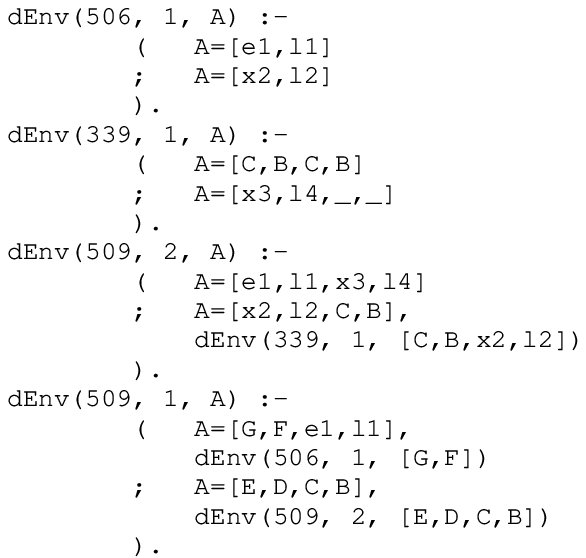}}
\hspace*{4ex}
\end{minipage}
\caption{Packed UDRS: conjunctive part (left column) and disjunctive
  binding environment}
\label{fig:output-udrs}
\end{figure*}

\section{Implementation and Experimental Results}
\label{sec:impl}

The algorithm has been implemented for the {\sc Prolog} (or DCG)
constraint system, i.e., constraints are equations over first-order
terms. Two implementations have been done. One in the concurrent
constraint language OZ \cite{oz-general96} and one in Sicstus Prolog.%
\footnote{The OZ implementation has the advantage that feature
  structure constraint solving is built-in. Our implementation
  actually represents the DCG terms as a feature structures.
  Unfortunately it is an order of magnitude slower than the Prolog
  version. The reason for this presumably lies in the fact that
  meta-logical operations the algorithm needs, like {\tt generalise}
  and {\tt copy\_term} have been modeled in OZ and not on the logical
  level were they properly belong, namely the constraint solver.}
The following results relate to the Prolog implementation.%
\footnote{This implementation is available from {\sf
    http://www.ims.uni-stuttgart.de/\~{}jochen/CBSem}.}

\begin{figure}[tb]
\centerline{\psfig{file=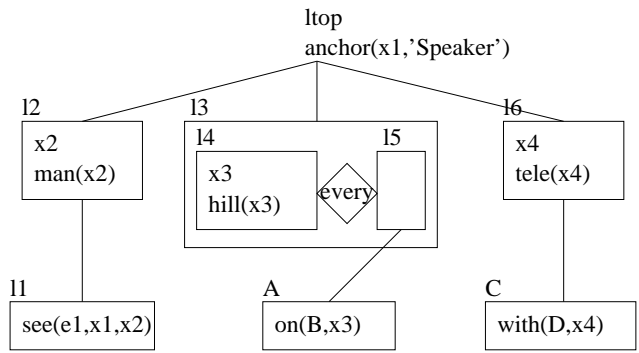}}
\caption{Conjunctive part of UDRS, graphically}
\label{fig:udrs-graph}
\end{figure}

Fig.~\ref{fig:output-udrs} shows the resulting packed UDRS 
for the example forest in Fig.~\ref{fig:forest1}.
Fig.~\ref{fig:udrs-graph} displays the
SEM part as a graph.
The disjunctive binding environment only
encodes what the variable referents $B$ and $D$ (in conjunction with
the corresponding labels $A$ and $C$) may be bound to
to: one of $e1$, $x2$, or $x3$ (and likewise the corresponding label).
Executing the goal {\tt dEnv(509,1,[B,A,D,C])} yields the five
solutions:
{\small
\begin{verbatim}
A = l1, B = e1, C = l1, D = e1 ? ;
A = l2, B = x2, C = l1, D = e1 ? ;
A = l1, B = e1, C = l4, D = x3 ? ;
A = l2, B = x2, C = l2, D = x2 ? ;
A = l2, B = x2, C = l4, D = x3 ? ;

no
| ?- 
\end{verbatim}
}

Table~\ref{tab:timing} gives execution times used for semantics
\begin{table}
\begin{center}
\vspace{-\abovedisplayskip}
\begin{tabular}{|l|l|l|l|}
\hline
$n$ & Readings & {\sc AND}- + {\sc OR}-nodes & Time\\
\hline
2 & 5 & 35 & 4 msec\\
4 & 42 & 91 & 16 msec\\
6 & 429 & 183 & 48 msec\\
8 & 4862 & 319 & 114 msec\\
10 & 58786 & 507 & 220 msec\\
12 & 742900 & 755 & 430 msec\\
14 & 9694845 & 1071 & 730 msec\\
16 & 129 Mio. & 1463 & 1140 msec\\
\hline
\end{tabular}
\end{center}
\caption{Execution times}
\label{tab:timing}
\end{table}
construction of sentences of the form {\sf I saw a man (on a
  hill)$^n$} for different $n$. The machine used for the experiment
was a Sun Ultra-2 (168MHz), running Sicstus 3.0\#3. In a further
  experiment an n-ary {\tt anti\_unify} operation was implemented,
  which improved execution times for the larger sentences, e.g., the
  16 PP sentence took 750 msec. These results
  approximately fit the expectations from the theoretical complexity
  bound.

\section{Discussion}
\label{sec:discussion}

Our algorithm and its implementation show that it is
not only possible in theory, but also feasible in practice to
construct packed semantical representations directly from parse
forests for sentence that exhibit massive syntactic ambiguity. The
algorithm is both in asymptotic complexity and in real numbers
dramatically faster than an earlier approach, that also tries to
provide an underspecified semantics for syntactic ambiguities. The
algorithm has been presented abstractly from the actual constraint
system and can be adapted to any constraint-based grammar formalism. 

A critical assumption for the method has been that semantic rules
never fail, i.e., no search is involved in semantics construction.
This is required to guarantee that the resulting constraint is a kind
of `solved form' actually representing so-to-speak the free
combination of choices it contains. Nevertheless, our method (modulo
small changes to handle failure) may still
prove useful, when this restriction is not fulfilled, since it focuses
on computing the common information of disjunctive branches. The
conjunctive part of the output constraint of the algorithm can then be
seen as an approximation of the actual result, if the output
constraint is satisfiable. Moreover, the disjunctive parts are
reduced, so that a subsequent full-fledged search will have
considerably less work than when directly trying to solve the original
constraint system.

\end{document}